\documentclass[11pt]{article}
\usepackage{mtsummit2019}
\usepackage{natbib}
\usepackage{times}
\usepackage{url}
\usepackage{hyperref}
\usepackage{latexsym}
\usepackage[small,bf]{caption} 
\usepackage{booktabs}
\usepackage{graphicx}
\usepackage{bm}
\usepackage{amsmath}
\usepackage{subcaption}
\usepackage{xcolor}

\newenvironment{itemizesquish}
{\begin{list}{\labelitemi}
{\setlength{\itemsep}{0em}\setlength{\labelwidth}
{0.5em}\setlength{\leftmargin}
{\labelwidth}\addtolength{\leftmargin}
{\labelsep}}}{\end{list}}

\title{Translator2Vec: Understanding and Representing Human Post-Editors}

\author{Ant\'onio G\'ois \smallskip \\ 
  Unbabel\\
  Lisbon, Portugal  \smallskip\\
  {\tt antonio.gois@unbabel.com}  \And
  Andr\'e F. T. Martins \smallskip \\
  Unbabel \& Instituto de Telecomunica\c{c}\~oes \\
  Lisbon, Portugal  \smallskip\\
  {\tt andre.martins@unbabel.com}}

\date{}

\begin{document}
\maketitle
\begin{abstract}
The combination of machines and humans for translation is effective, with many studies showing productivity gains when humans post-edit machine-translated output instead of translating from scratch. 
To take full advantage of this combination, we need a fine-grained understanding of how human translators work, and which post-editing styles are more effective than others. 
In this paper, we release and analyze a new dataset with document-level post-editing action sequences, including edit operations from keystrokes, mouse actions, and waiting times. Our dataset comprises 66,268 full document sessions post-edited by 332 humans, the largest of the kind released to date. We show that action sequences are informative enough to identify post-editors accurately, compared to baselines that only look at the initial and final text. We build on this to learn and visualize continuous representations of post-editors, and we show that these representations improve the downstream task of predicting post-editing time.
\end{abstract}

\section{Introduction}\label{sec:intro}

Computer-aided translation platforms for interactive translation and post-editing are now commonly used in professional translation services \citep{Alabau2014,Federico2014,Green2014,Denkowski2015,Hokamp2018, sin2014development, kenny2011electronic}. 
With the increasing quality of machine translation \citep{Bahdanau2014,Gehring2017,Vaswani2017}, the translation industry is going through a transformation, progressively shifting gears from ``computer-aided'' (where MT is used as an instrument to help professional translators) towards {\bf human-aided translation}, where there is a human in the loop who only intervenes when needed to ensure final quality, and whose productivity is to be optimized. 
A deep, data-driven understanding of the {\bf human post-editing process} is key to achieve the best trade-offs in translation efficiency and quality. What makes a ``good'' post-editor? What kind of behaviour shall an interface promote?


There is a string of prior work that relates the difficulty of translating text with the cognitive load of human translators and post-editors, based on indicators such as editing times, pauses, keystroke logs, and eye tracking \citep[see also \S\ref{sec:related}]{OBrien2006,Doherty2010,LaCruz2012,Balling2014}. Most of these studies, however, have been performed in controlled environments on a very small scale, with a limited number of professional translators and only a few sessions. A direct use of human activity data for understanding and representing human post-editors, towards improving their productivity, is still missing, arguably due to the lack of large-scale data. Understanding how human post-editors work could open the door to the design of better interfaces, smarter allocation of human translators to content, and automatic post-editing.

In this paper, 
we study the behaviour of human post-editors ``in the wild'' by automatically examining tens of thousands of post-editing sessions at a document level. 
We show that these detailed editor activities (which we call {\bf action sequences}, \S\ref{sec:actions}) encode useful additional information besides just the initial machine-translated text and the final post-edited text.  
This is aligned to recent findings in other domains: \citet{yang2017identifying_semantic_intent} and \citet{faruqui2018wiki_edits} have recently shown that Wikipedia page edits 
can represent interesting linguistic phenomena in language modeling and discourse. 
While prior work analyzed the cognitive behaviour of post-editors and their productivity by collecting a few statistics, we take a step forward in this paper, using state-of-the-art machine learning techniques to {\bf represent editors in a vector space} (\S\ref{sec:editor_representation}). These representations are obtained by training a model to {\bf identify} the editor based on his action sequences (\S\ref{sec:editor_identification}). This model achieves high accuracy in predicting the editor's identity, and the learned representations exhibit interesting correlations with the editors' behaviour and their productivity, being effective when plugged as features for {\bf predicting the post-editing time} (\S\ref{sec:downstream_task}).

Overall, we use our action sequence dataset  to address the following research questions:
\begin{enumerate}
\item {\bf Editor identification} (\S\ref{sec:editor_identification}): are the post-editors' activities (their action sequences) informative enough to allow discriminating their identities from one another (compared to just using the initial machine-translated text and the final post-edited one)?
\item {\bf Editor representation} (\S\ref{sec:editor_representation}): can the post-editors' activities be used to learn meaningful vector representations, such that similar editors are clustered together? Can we interpret these embeddings to understand which activity patterns characterize ``good'' editors (in terms of translation quality and speed)?
\item {\bf Downstream tasks} (\S\ref{sec:downstream_task}): do the learned editor vector representations provide useful information for downstream tasks, such as predicting the time to translate a document, compared to pure text-based approaches that do not use them?
\end{enumerate}
We base our study on editor-labeled action sequences for two language pairs, English-French and English-German, which we make available for future research. In both cases, we obtain positive answers to the three questions above.

\section{Post-Editor Action Sequences}\label{sec:actions}

\begin{table}[t]
\centering
\small
\begin{tabular}{lll}
\toprule
Action       & Symbol & Appended Info       \\
\midrule 
Replace      & {\tt R}             & new word           \\ 
Insert       & {\tt I}             & new word           \\ 
Delete       & {\tt D}             & old word           \\ 
Insert Block & {\tt BI}            & new block of words \\ 
Delete Block & {\tt BD}            & old block of words \\
\midrule
Jump Forward             & {\tt JF}             & \# words          \\ 
Jump Back         & {\tt JB}            & \# words          \\ 
Jump Sentence Forward      & {\tt JSF}            & \# sentences         \\ 
Jump Sentence Back & {\tt JSB}           & \# sentences         \\ 
Mouse Clicks      & {\tt MC}            & \# mouse clicks     \\ 
Mouse Selections  & {\tt MS}            & \# mouse selections \\ 
Wait              & {\tt W}             & time (seconds)                   \\ 
Stop              & {\tt S}             & --                      \\ 
\bottomrule
\end{tabular}
\caption{Text-editing and non-editing actions.}
\label{tab:action_types}
\end{table}

A crucial part of our work is in converting raw keystroke sequences and timestamps into {\bf action sequences}---sequences of symbols in a finite alphabet that describe word edit operations (insertions, deletions, and replacements), batch operations (cutting and pasting text), mouse clicks or selections, jump movements, and pauses. 



Each action sequence corresponds to a single post-editing session, in which a human post-edits a document. The starting point is a set of source documents (customer service email messages), which are sent for translation to Unbabel's online translation service. The documents are split into sentences and translated by a domain-adapted neural machine translation system based on Marian \citep{JunczysDowmunt2018}. 
Finally, each document is assigned to a human post-editor to correct eventual translation mistakes.%
\footnote{The human post-editors are native or proficient speakers of both source and target languages, although not necessarily professional translators. They are evaluated on language skills and subject to periodic evaluations by Unbabel. Editors have access to whole documents when translating, and they are given content-specific guidelines, including style, register, etc.} %
These post-editing sessions are logged, and all the keystroke and mouse operation events are saved, along with timestamps. 
A preprocessing script converts these raw keystrokes into word-level action sequences, as we next describe, and a unique identifier is appended that represents the human editor. %



\begin{table*}[]
\centering
\small
\begin{tabular}{ll}
\toprule
Source  & \begin{tabular}[c]{@{}l@{}}Hey there,\\ Some agents do speak Spanish, otherwise our system will translate :)\\ Best,\\ \textless{}Name\textgreater{}\end{tabular}           \\
\midrule
MT      & \begin{tabular}[c]{@{}l@{}}Bonjour,\\ Certains agents parlent espagnol, sinon notre syst\`eme \textcolor{red}{\it se traduira par} :)\\ Cordialement,\\ \textless{}Name\textgreater{}\end{tabular} \\
\midrule
PE      & \begin{tabular}[c]{@{}l@{}}Bonjour,\\ Certains agents parlent espagnol, sinon notre syst\`eme \textcolor{blue}{\bf traduit} :)\\ Cordialement,\\ \textless{}Name\textgreater{}\end{tabular}         \\
\midrule
Actions & \begin{tabular}[c]{@{}l@{}}{\tt W}:23  \quad  {\tt JSF}:1  \quad  {\tt JF}:8  \quad  {\tt D}:se  \quad  {\tt W}:2 \quad   {\tt MC}:1  \quad  {\tt MS}:1  \quad  {\tt JF}:1  \quad  {\tt D}:par \quad {\tt W}:7 \\  {\tt MC}:1  \quad  {\tt MS}:1  \quad  {\tt JB}:1 \quad  {\tt R}:traduit \quad  {\tt W}:2 \quad {\tt MS}:1 \quad   {\tt S}:-- \end{tabular}\\
\bottomrule
\end{tabular}
\caption{Example of a document and corresponding action sequence. We mark in \textcolor{red}{\it red} the MT words that have been corrected and in \textcolor{blue}{\bf blue} their replacement. The actions used here were {\tt W} (wait), {\tt JSF} (jump sentence forward), {\tt JF} (jump forward), {\tt D} (delete), {\tt MC} (mouse clicks), {\tt MS} (mouse selections), {\tt JB} (jump back), {\tt R} (replace) and {\tt S} (stop).}
\label{table:sample_sequences}
\end{table*}

The preprocessing for converting the raw character-level keystroke data into word-level actions is as follows. We begin with a sequence of all intermediate states of a document between the machine-translated and the post-edited text, containing changes caused by each keystroke. We track the position of the word currently being edited and store one action summarizing the change in that word. A single keystroke may also cause simultaneous changes to several words (e.g. when pasting text or deleting a selected block), and we reserve separate actions for these. Overall, five {\bf text-editing actions} are considered: inserting ({\tt I}), deleting ({\tt D}), and replacing ({\tt R}) a single word, and inserting ({\tt BI}) and deleting ({\tt BD}) a block of words. Each action is appended with the corresponding word or block of words, as shown in Table \ref{tab:action_types}.


Other actions, dubbed {\bf non-editing actions}, do not change the text directly. Jump-forward ({\tt JF}) and jump-backward operations ({\tt JB}) count the distance in words between two consecutive edits. 
Another pair of actions informs when a new sentence is edited: a sentence jump ({\tt JSF}/{\tt JSB}) indicates that we moved a certain number of sentences forth/back since the previous edit. 
Mouse clicks ({\tt MC}) and mouse selections ({\tt MS}) count their occurrences between two consecutive edits. Wait ({\tt W}) counts the seconds between the beginning of two consecutive edits. Finally, stop ({\tt S}) marks the end of the post-editing session.


Since we do not want to rely on lexical information to identify the human post-editors, only the 50 most frequent words were kept (most containing punctuation symbols and  stop-words), with the remaining ones converted to a special unknown symbol ({\tt UNK}). 
Moreover, the first waiting time is split in two: the time until the first keystroke occurs and, in case the first keystroke is not part of the first action (e.g. a mouse click), a second waiting time until the first action begins.


Table~\ref{table:sample_sequences} shows an example of a small document, along with the editor's action sequence. The editor began on sentence 2 (``Certains agents...") and the word on position 9, since there was a jump forward of 1 sentence and 8 words. After deleting ``se", position 9 became ``traduira". Since the editor opted to delete ``par" (using a mouse selection) before changing the verb, there is a jump forward of 1 word to position 10. Then we have a jump back of 1 before changing the verb to ``traduit".


\paragraph{Datasets.} 
We introduce two datasets for this task, one for English-French (En-Fr) and another for English-German (En-De). For each dataset, we provide the action sequences for full documents, along with an editor identifier. 
To ensure reproducibility of our results, we release both datasets as part of this paper, available in \url{https://github.com/Unbabel/translator2vec/releases/download/v1.0/keystrokes_dataset.zip}. 
For anonymization purposes, we convert all editor names and the 50 tokens in the word vocabulary to numeric identifiers. 
Statistics of the dataset are shown in Table~\ref{table:datasets}: it is the largest ever released dataset with post-editing action sequences, and the only one we are aware of with document-level information.%
\footnote{The closest comparable dataset was released by \citet{Specia2017} in the scope of the QT21 project, containing 176,476 sentences spanning multiple language pairs (about 4 times less), with raw keystroke sequences being available by request. In contrast to ours, their units are sentences and not full documents, which precludes studying how human post-editors jump between sentences when translating a document.} Each document corresponds to a customer service email with an average of 116.6 tokens per document. Each sentence has an average length of 9.4 tokens.


\begin{table}[]
\begin{tabular}{llrrr}
\toprule
&
& \# docs 
& \# sents 
& \# words
\\
\midrule
& train 
& {17,464}
& {154,026}
& {1,895,389} 
\\
En-Fr
& dev
& {5,514}
& {52,366}
& {659,675}
\\
& test
& {9,441}
& {86,111}
& {1,072,807}
\\
\midrule
& train
& {17,403}
& {169,478}
& {2,053,407}
\\
En-De
& dev
& {6,722}
& {66,521}
& {826,791}
\\
& test
& {9,724}
& {98,920}
& {1,221,319}
\\
\midrule
Total && 66,268 & 627,422 & 7,729,388\\
\bottomrule
\end{tabular}
\caption{Number of documents, sentences, and words in English source text per dataset. There are 149 unique editors across all En-Fr datasets, and 183 in En-De.}
\label{table:datasets}
\end{table}


\begin{figure*}[htb!]
    \centering
    \includegraphics[width=\textwidth]{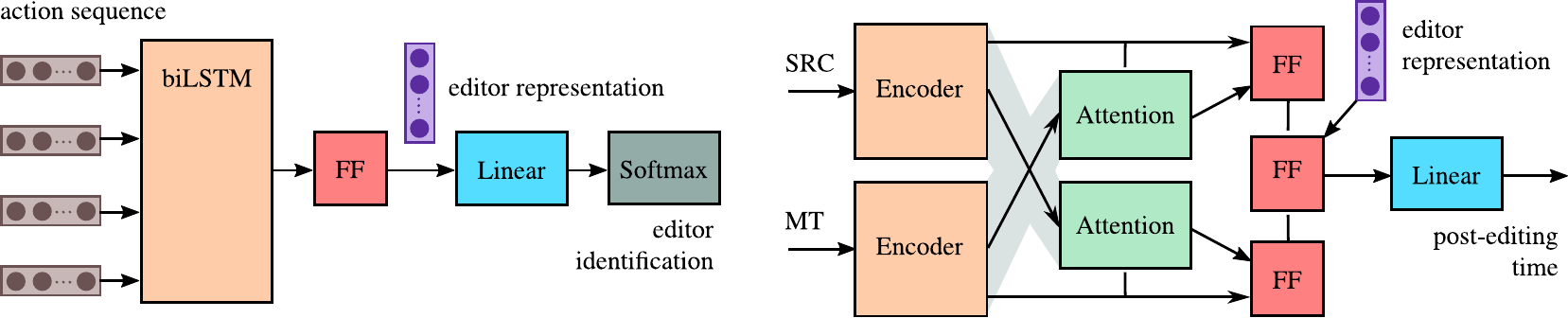}
    \caption{Left: our {\bf Action Seq} model for editor identification. Right: our model for post-editing time prediction.
    }
    \label{fig:lstm}
\end{figure*}

\section{Editor Identification} \label{sec:editor_identification}



We now make use of the dataset just described to answer the three research questions stated at the end of \S\ref{sec:intro}, starting with {\bf editor identification}. 
\subsection{Data Preparation}

For this experiment, we took the action sequence dataset described in \S\ref{sec:actions} and selected a small number of human translators for both language pairs who post-edited a number of documents above a threshold: this yielded 6 editors for En-Fr and 7 editors for En-De. To ensure balanced datasets, we filtered them to contain the same number of samples per selected editor. This filtering yielded a total of 998/58/58 training/dev/test documents per editor for En-Fr, and 641/128/72 for En-De. 

A random baseline for this dataset would obtain an editor identification accuracy of $1/6 = 16.7\%$ for En-Fr and $1/7 =14.3\%$ for En-De.


\subsection{A Model for Editor Identification}\label{sec:model}


Let $\langle x_1, \ldots, x_L \rangle$ be an action sequence produced by a post-editor $y$. To identify the editor of a task, we build a model $P(y \mid x_1, \ldots, x_L)$ using a neural network as we next describe (shown in Figure~\ref{fig:lstm}). 
Each action $x_i$ is first associated to a one-hot vector. All numeric actions are grouped into bins---e.g. waiting times of 200 seconds and higher all correspond to the same one-hot representation. Bins were defined manually, providing higher granularity to small values than to larger ones.%
\footnote{We used $\{0, \ldots, 5, 7, 10, 15, 20, 30, 50, 75, 100, 150, 200+\}$ for wait and jump events (in seconds and word positions, respectively); and $\{0, \ldots, 5, 7, 10+\}$ for sentence jumps and mouse events (in sentence positions and clicks).} %
Each one-hot is then mapped to a learnable embedding, and the sequence of embeddings is fed to a 2-layer bidirectional LSTM (biLSTM; \citet{Hochreiter1997,Graves2005}), resulting in two final states $\bm{\overrightarrow{h}}, \bm{\overleftarrow{h}}$. Then we concatenate both, apply dropout \citep{Srivastava2014} and feed them to a feed-forward layer with a ReLU activation \citep{Glorot2011} to form a vector $\bm{h}$. 
This vector is taken as the representation of the action sequence. 
Finally, we define $P(y \mid x_1, \ldots, x_L) = \mathsf{softmax}(\bm{W}\bm{h} + \bm{b})$.

We call this model {\bf Action Seq}, since it exploits information from the action sequences.

\subsection{Baselines}\label{sec:baselines}

To assess how much information action sequences provide about human editors beyond the initial (machine translation) and final (post-edited) text, we implemented various baselines which do \textbf{not} use fine-grained information from the action sequences. All use pre-trained text embeddings from FastText \citep{Joulin2017}, and they are all tuned for dropout and learning rate:
\begin{itemizesquish}
    \item One using  the machine-translated text only ({\bf MT}). Since this text has not been touched by the human post-editor, we expect this system to perform similarly to the random baseline. The goal of this baseline is to control whether there is a bias in the content each editor receives that could discriminate her identity. It uses word embeddings as input to a biLSTM, followed by feed-forward and softmax layers. 
    \item Another one using  the posted-edited text only ({\bf PE}). This is used to control for the linguistic style of the post-editor. We expect this to be a weak baseline, since although there are positive results on translator stylometry \citep{translator_stylometry_elfiqi2019}, the task of post-editing provides less opportunity to leave a fingerprint than if writing a translation from scratch. The architecture is the same as in the {\bf MT} baseline.
    \item A baseline combining both MT and PE using a dual encoder architecture ({\bf MT + PE}), inspired by models from dialogue response \citep{lowe2016de, lu2017dialogueresponse}. This baseline is stronger than the previous two, since it is able to look at the {\it differences} between the initial and final text produced by the post-editor, although it ignores the process by which these differences have been generated. Two separate biLSTMs encode the two sequences of word embeddings, the final encoded states are concatenated and fed to a feed-forward and a softmax layer to provide the editors' probabilities.
    \item Finally, a stronger baseline ({\bf MT + PE + Att}) that is able to ``align'' the MT and PE, by augmenting the dual encoder above with an attention mechanism, inspired by work in natural language inference \citep{Rocktaschel2016}. The model resembles the one in Figure~\ref{fig:lstm} (right), with a softmax output layer and without the editor representation layer. Two separate biLSTMs are used to encode the machine-translated and the post-edited text. The final state of the MT is used to compute attention over the PE, then this attention-weighted PE is concatenated with MT's final state and passed through a feed-forward layer. Symmetrically we obtain a representation from PE's final state and an attention-weighted MT. Finally both vectors are concatenated and turned into editors' probabilities through another feed-forward layer.
\end{itemizesquish}

\begin{table}[t]
\centering
\begin{tabular}{lcc}
\toprule
& En-De (\%) & En-Fr (\%)\\
\midrule
{\bf Delta} & 16.15 & 26.09 \\ 
{\bf MT} & 18.21 & 16.44\\
{\bf PE} & 27.38 & 30.00\\
{\bf MT + PE} & 26.63 & 31.78\\
{\bf MT + PE + Att} & 30.12 & 35.06\\ 
\midrule
{\bf Action Seq} & {\bf 84.37} & {\bf 67.07} \\
\bottomrule
\end{tabular}
\caption{Results of model and baselines for editor identification. Reported are average test set accuracies of 5 runs, with 7 editors for En-De and 6 editors for En-Fr.}
\label{table:results}
\end{table}

Additionally, we prepare another baseline ({\bf Delta}) as a tuple with meta information containing statistics about the difference between the initial and final text (still not depending on the action sequences). This tuple contains the following 5 elements: a count of sentences in the document, minimum edit distance between MT and PE, count of words in the original document, in MT and in PE. Each of these elements is binned and mapped to a learnable embedding. 
The 5 embeddings are concatenated into a vector $\bm{e}$, followed by a feedforward layer and a softmax activation. 

\begin{table}[t]
\centering
\begin{tabular}{lcc}
\toprule
& En-De (\%) & En-Fr (\%)\\
\midrule
{\bf Action Seq} & {\bf 83.31} & {\bf 73.16} \\
\midrule
w/out editing actions & 80.60 & 69.37\\
w/out mouse info & 75.49 & 66.38\\
w/out waiting time & 80.42 & 70.92\\
w/out 1st waiting time & 78.60 & 71.15\\
only editing actions & 60.20 & 59.08\\
only mouse info  & 56.43 & 55.06\\                   only waiting time & 53.53 & 44.02\\
only 1st waiting time & 24.22 & 23.11\\
\bottomrule
\end{tabular}
\caption{Ablations studies for editor identification. Reported are average development set accuracies of 5 runs, with 7 editors for En-De and 6 editors for En-Fr.}
\label{table:ablations}
\end{table}

\subsection{Editor Identification Accuracy}

Table~\ref{table:results} compares our system with the baselines above. 
Among the baselines, we observe a gradual improvement as models have access to more information. The fact that the MT baseline performs closely to the random baseline is reassuring, showing that there is no bias in the type of text that each editor receives. As expected, the dual encoder model with attention, being able to attend to each word of the MT and post-edited text, is the one which performs the best, surpassing the random baseline by a large margin. However, none of these baselines have a satisfactory performance on the editor identification task. 

By contrast, 
the accuracies achieved by our proposed model ({\bf Action Seq}) are striking: 84.37\% in En-De and 67.07\% in En-Fr, way above the closest baselines. 
This large gap confirms our hypothesis that {\bf the editing process itself contains information which is much richer than the initial and final text only}. 

\paragraph{Ablation studies.} 
To understand the importance of each action type in predicting the editor's identity, we conduct a series of ablation studies and report development set accuracies in Table~\ref{table:ablations}. These experiments involve removing mouse information, time information, initial waiting time or editing actions. Also, we try keeping only each of the previous four. We find that all action types contribute to the global accuracy, although to different extents. Also, some action types achieve high performance on their own. Somewhat surprisingly, mouse information alone achieves remarkably high accuracy. Although waiting times also perform well on their own, removing them has little impact on the final score.

\section{Editor Representation}\label{sec:editor_representation}

The previous section has shown how the action sequences are very effective for identifying editors. As a by-product, the {\bf Action Seq} model used for that task produced an internal vector $\boldsymbol{h}$ that represents the full post-editing session. This suggests a strategy for obtaining {\bf editor representations}: simply \textit{average} all such vectors from each editor.  
One way of looking at this is regarding editor identification as an auxiliary task that assists us in finding good editor representations. This draws inspiration from previous work, such as \citet{Mikolov2013}, as well as its applications to recommendation systems \citep{yahoo_ecommerce_inbox_Grbovic_2015, yahoo_queries_ads_grbovic2016}. In the last two works, an auxiliary task also helps to provide a latent representation of an object of interest.

\begin{figure}[htb!]
    \centering
    \begin{subfigure}{.5\columnwidth}
        \centering
        \includegraphics[width=\columnwidth]{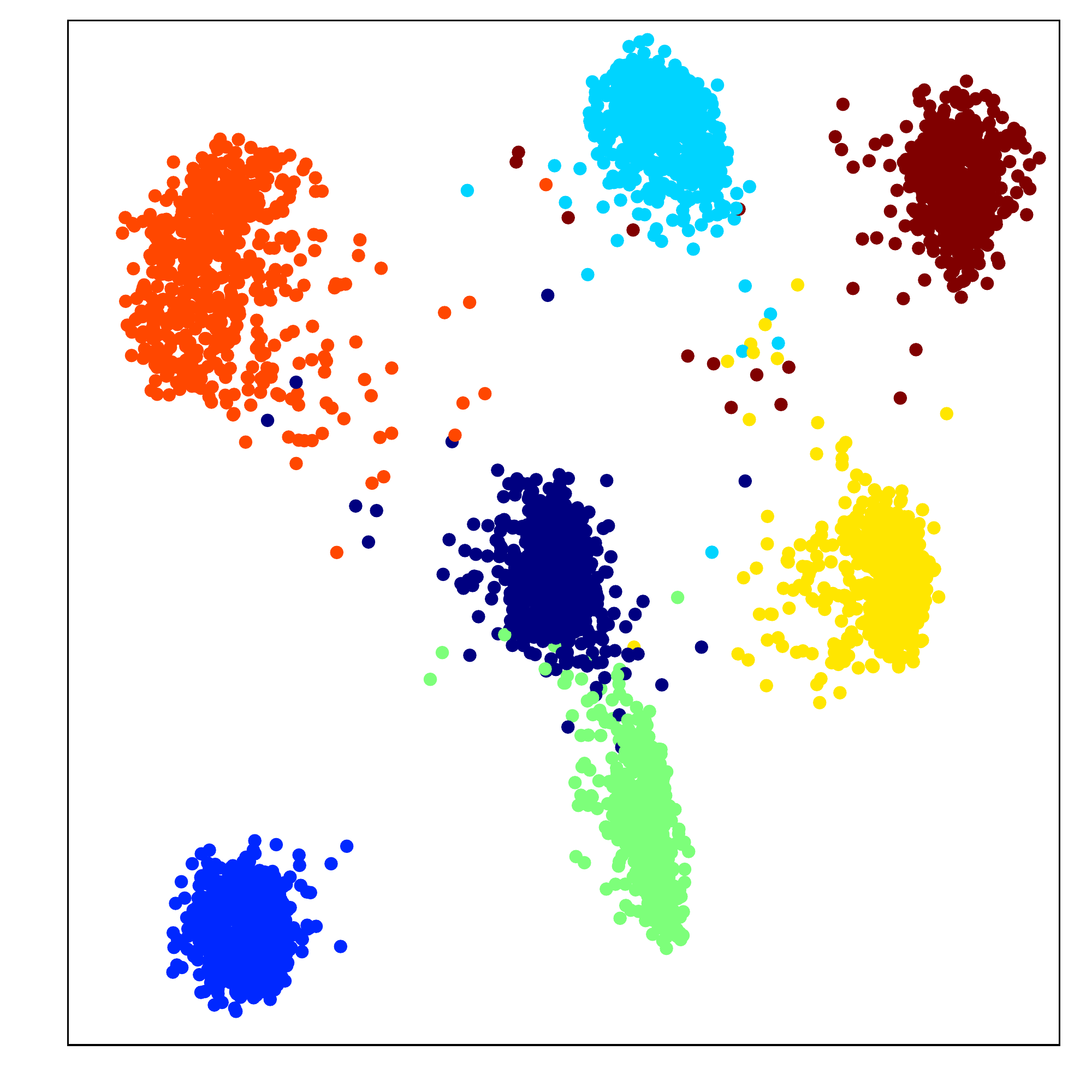}
        \caption{En-De training set}
        \label{fig:de_7eds_train}
    \end{subfigure}%
    \begin{subfigure}{.5\columnwidth}
        \centering
        \includegraphics[width=\columnwidth]{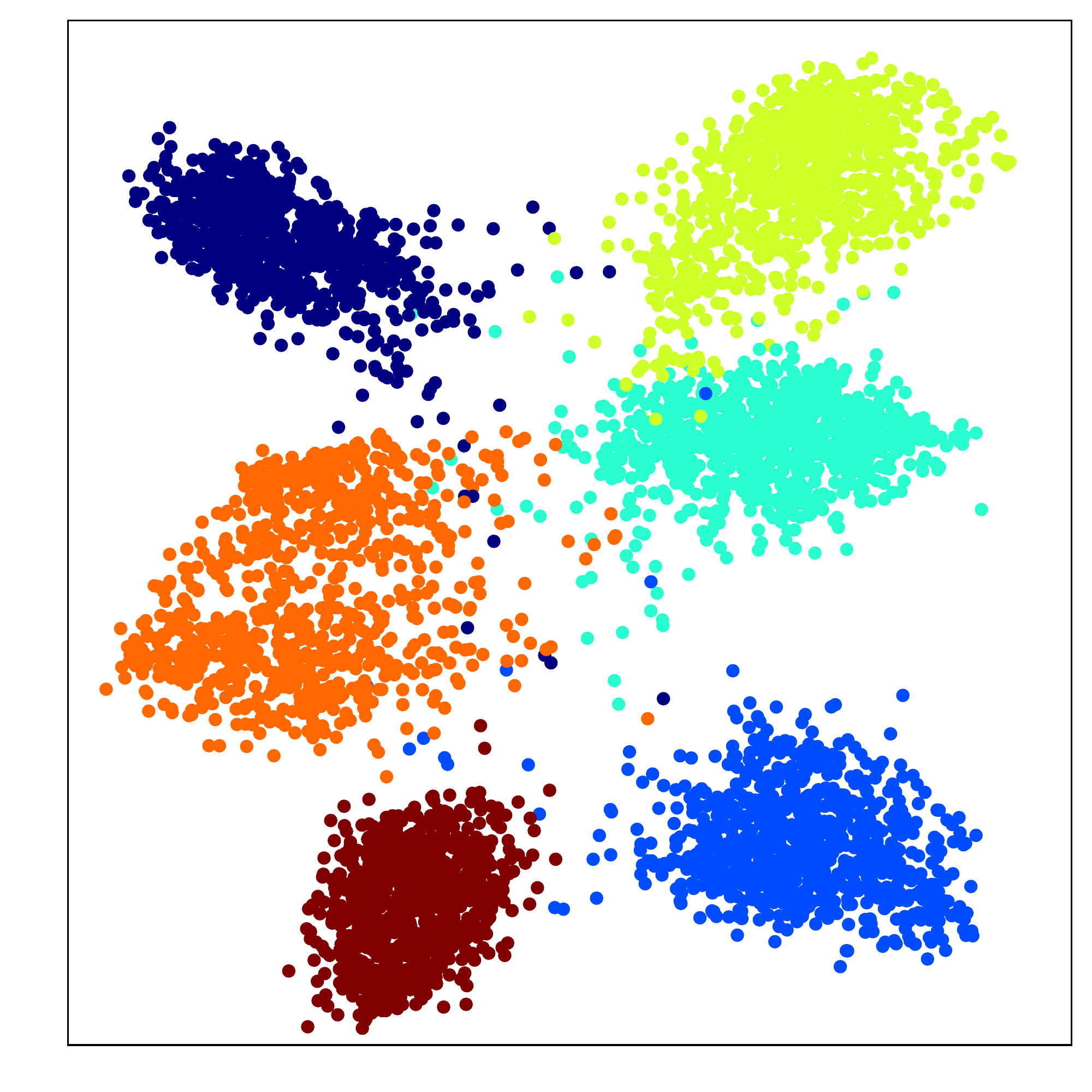}
        \caption{En-Fr training set}
        \label{fig:fr_6eds_train}
    \end{subfigure}
    \begin{subfigure}{.5\columnwidth}
        \centering
        \includegraphics[width=\columnwidth]{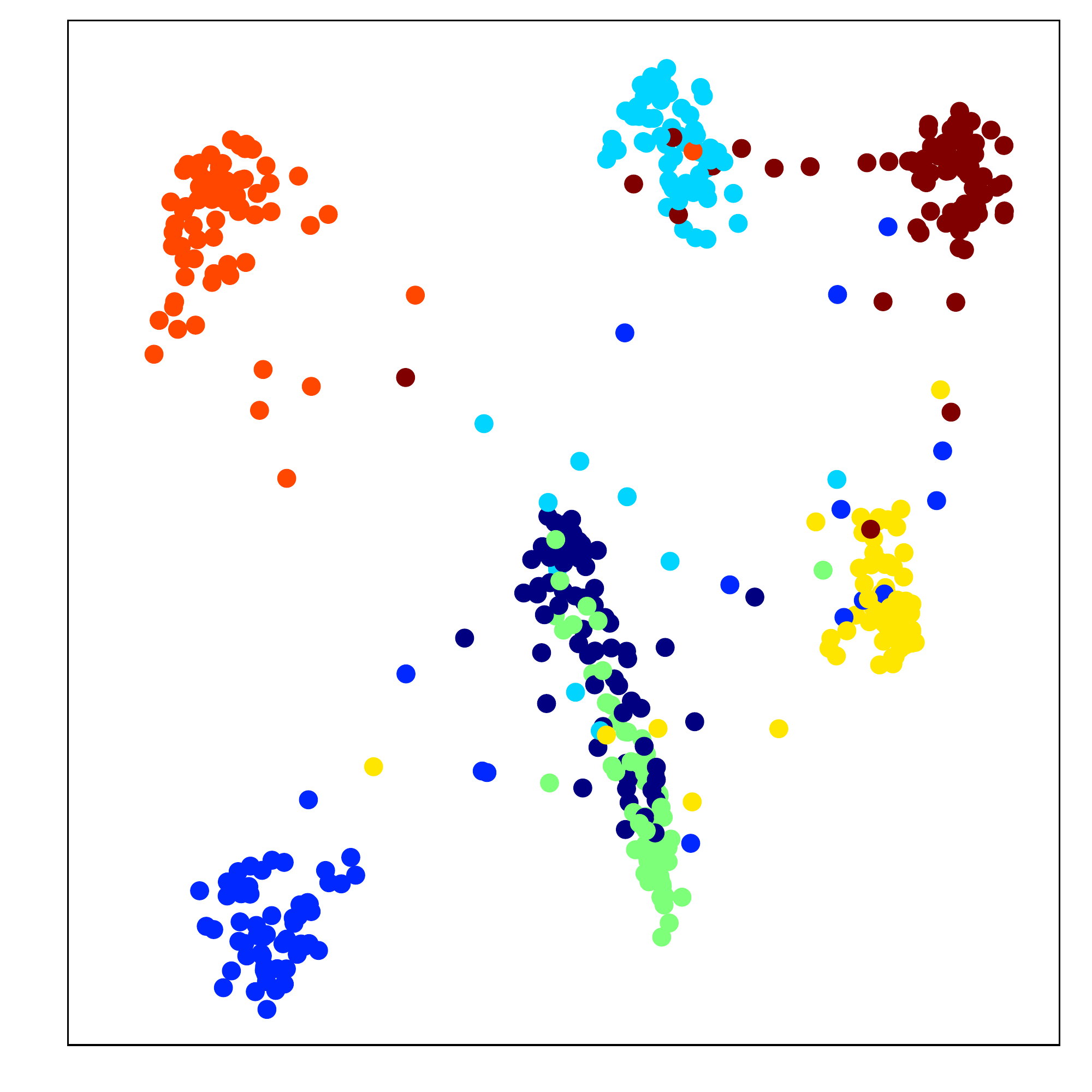}
        \caption{En-De test set}
        \label{fig:de_7eds_test}
    \end{subfigure}%
    \begin{subfigure}{.5\columnwidth}
        \centering
        \includegraphics[width=\columnwidth]{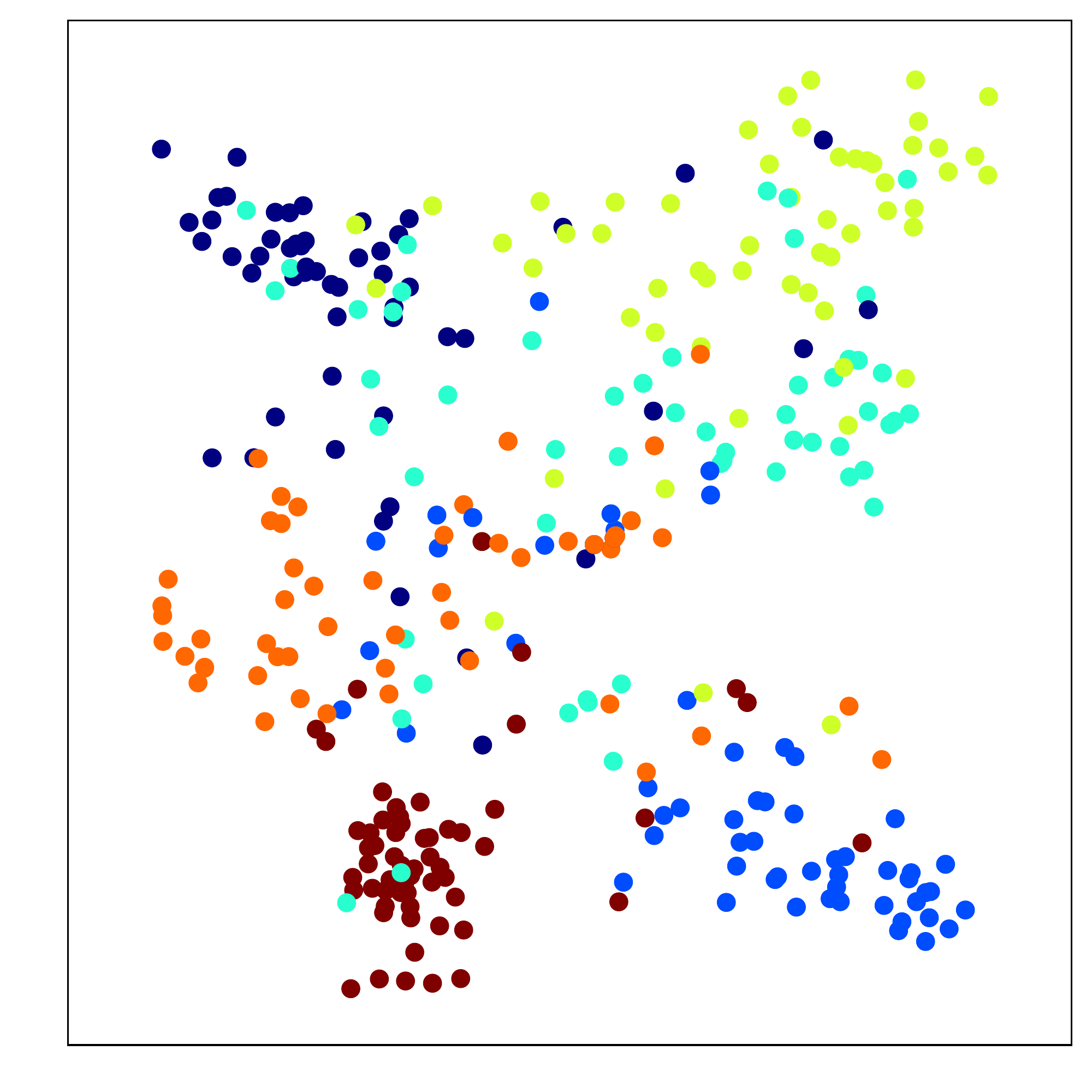}
        \caption{En-Fr test set}
        \label{fig:fr_6eds_test}
    \end{subfigure}
    \caption{Embeddings of each translation session in the editor identification train and test sets, with editors identified by different colors. For each language, the dimensionality reduction was learned by training parametric t-SNE \citep{parametricmaaten2009} on the train data, and then applying it to both train and test data. En-De contains 7 editors, each with 641 train and 72 test samples per editor. En-Fr contains 6 editors, each with 998 train and 58 test samples per editor.}
    \label{fig:sessions_tsne}
\end{figure}

\paragraph{Visualization of translation sessions.} 
To visualize the vectors $\bm{h}$ produced during our auxiliary task, we use Parametric t-SNE \citep{parametricmaaten2009} for dimensionality reduction. Unlike the original t-SNE \citep{tsnemaaten2008}, the parametric version allows to reapply a learned dimensionality reduction to new data. This way it is possible to infer a 2D structure using the training data, and check how well it fits the test data. 

In Figure~\ref{fig:sessions_tsne} we show a projection of vectors $\bm{h}$ for both language pairs, using a t-SNE model learned on the training set vectors; each color corresponds to a different editor. In the training set (used to train both the editor identification model and the Parametric t-SNE) there is one clear cluster for each editor, in both languages. Using test set data, new tasks also form clusters which are closely related to the editors' identity. Some clusters are isolated while others get mixed near their borders, possibly meaning that some editors behave in a more distinctive manner than others.

\paragraph{Visualization of editor representations.}

\begin{figure}[htb!]
    \centering
    \begin{subfigure}{.5\columnwidth}
        \centering
        \includegraphics[width=\columnwidth]{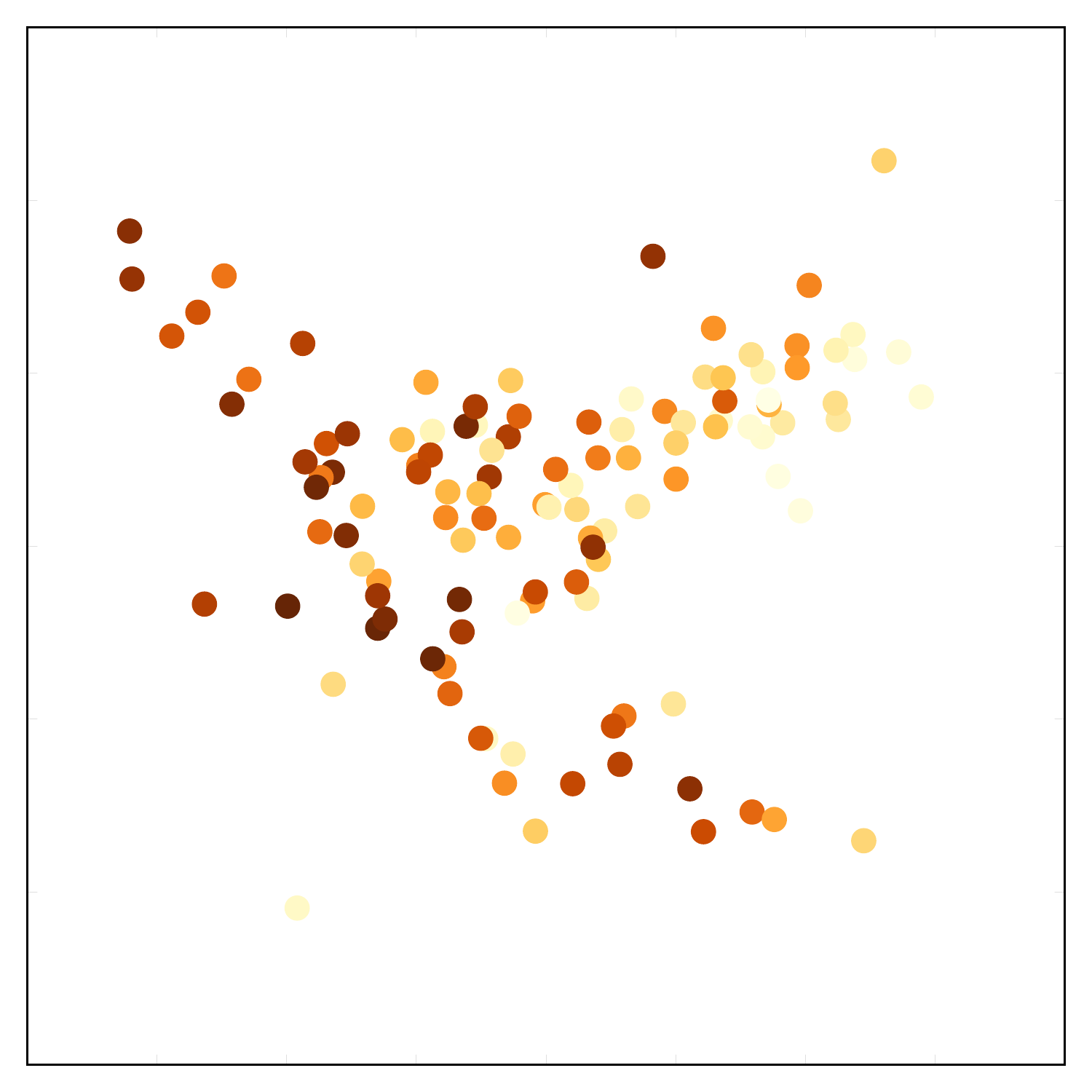}
        \caption{First wait time}
        \label{fig:editor_tsnes_firstW}
    \end{subfigure}%
    \begin{subfigure}{.5\columnwidth}
        \centering
        \includegraphics[width=\columnwidth]{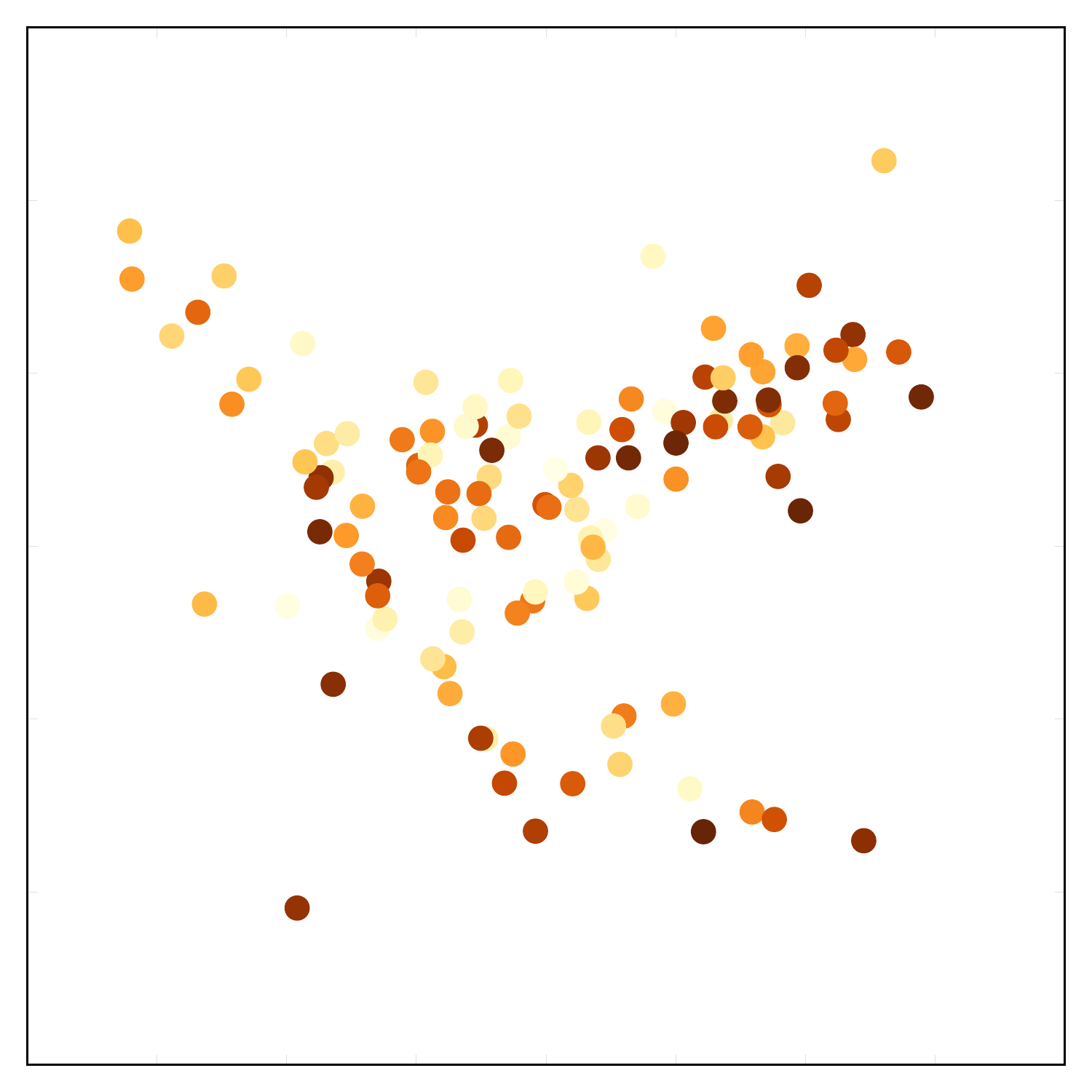}
        \caption{Jump-Backs}
        \label{fig:editor_tsnes_jbs}
    \end{subfigure}
    \begin{subfigure}{.61\columnwidth}
        \centering
        \includegraphics[width=\columnwidth]{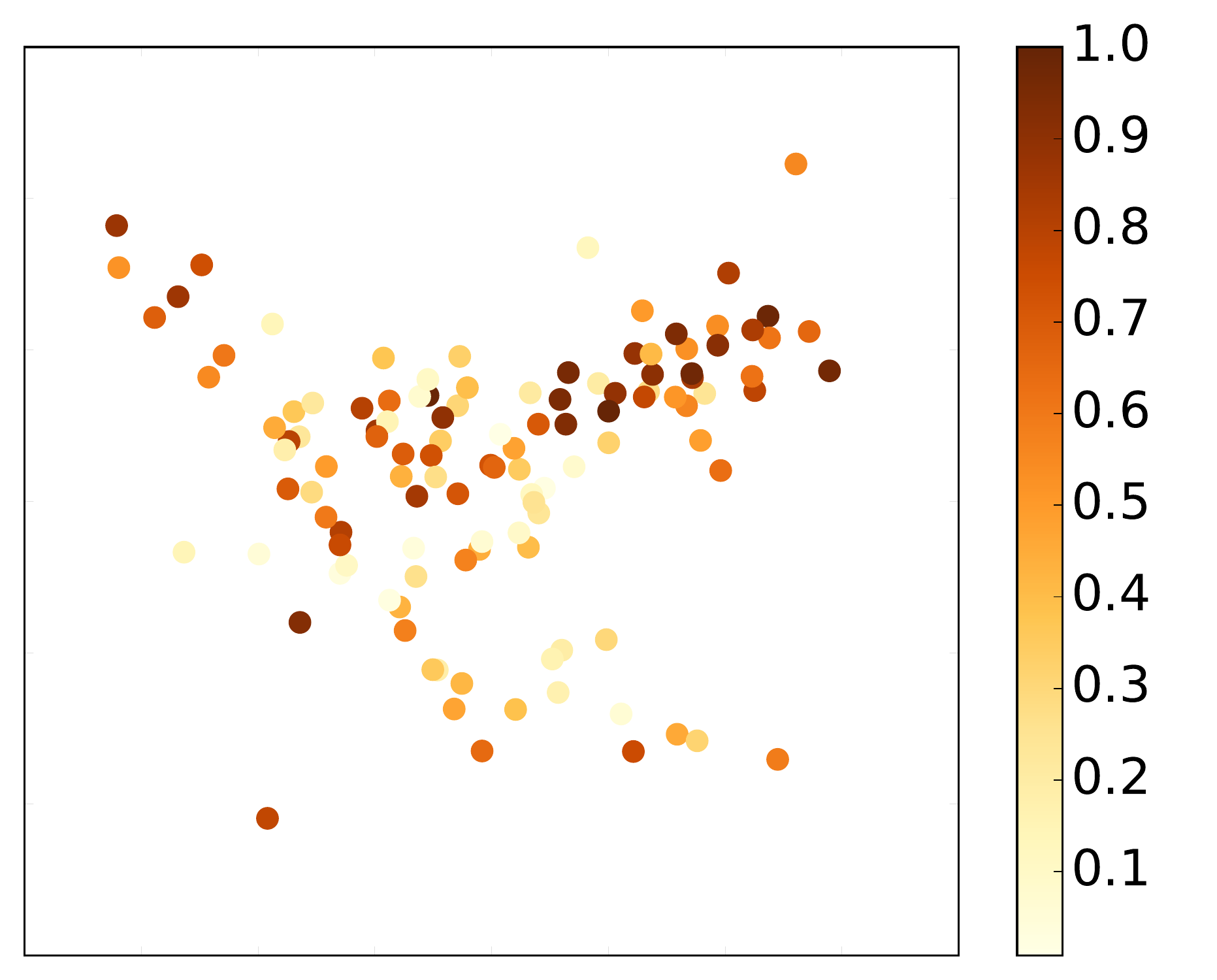}
        \caption{Mouse Events}
        \label{fig:editor_tsnes_mcms}
    \end{subfigure}%
    \caption{Embeddings of each En-Fr editor, mapped using the same parametric t-SNE as in Figure \ref{fig:sessions_tsne}. In \ref{fig:editor_tsnes_firstW} we have average pause before beginning for each editor, in percentile. In \ref{fig:editor_tsnes_jbs} we have the count of jump-backs per MT token of each editor, also in percentile. In \ref{fig:editor_tsnes_mcms} we have percentiles of counts of mouse clicks and selections per editor.}
    \label{fig:editor_tsnes}
\end{figure}

To represent an editor with a single vector, we average the $\bm{h}$'s of all tasks of a given editor to obtain his representation. Figure~\ref{fig:editor_tsnes} contains representations for En-Fr editors (similar results have been achieved for En-De editors), using the exact same model as in Figure~\ref{fig:fr_6eds_train} to produce session embeddings, and the same t-SNE model for visualization. To reduce noise we discard editors with less than 10 samples, keeping 117 out of 149 editors. 
In Figure~\ref{fig:editor_tsnes} we show percentiles for 3 editor features, using one point per editor and setting color to represent a different feature in each panel. In Figure~\ref{fig:editor_tsnes_firstW}, color represents percentiles of average initial waiting time, and in Figure~\ref{fig:editor_tsnes_jbs}, percentiles of counts of jump-backs per MT token. We can observe that the model learned to map high waiting times to the left and high counts of jump-backs to the right. In Figure~\ref{fig:editor_tsnes_mcms} we have mouse activity per user (percentiles of counts of mouse clicks and selections). Here we can see a distribution very similar to that of count of jump-backs. 


\begin{table}[]
\small
\centering
\begin{tabular}{lcc}
\toprule
   & Mouse and JB (\%) & 1st WT and JB (\%) \\ \midrule
En-Fr    & 80.75       & $-$39.65    \\
En-De    & 59.62       & $-$31.11    \\
\bottomrule
\end{tabular}
\caption{Pearson correlation between two pairs of variables: mouse actions / jump backs and first waiting time / jump backs.}
\label{table:correlations}
\end{table}

We hypothesize that there are two types of human editors: those who first read the full document and then post-edit it left to right; and those who read as they type, and go back and forth. 
To check these hypothesis, we measure the Pearson correlation between two pairs of variables in Table \ref{table:correlations}. Indeed, there is a slight negative correlation between the average initial pause and the count of jump backs per word. This  matches intuition, since a person who waited longer before beginning a task will probably have a clearer idea of what needs to be done in the first place. We also present the correlation between the count of mouse events (clicks and selections) and count of jump backs, which we observe to be very high. This may be due to the need to move between distant positions of the document, which is more commonly done with the mouse than with the keyboard.


\section{Prediction of Post-Editing Time}\label{sec:downstream_task}   

Finally, we design a downstream task with the goal of assessing the information contained in each translator's vector $\bm{h}$ and observing its applicability in a real-world setting. The task consists in predicting the post-editing time of a given job, which has been used as a quality estimation task in previous work \citep{logtimeprediction2013cohn, Specia2011}. As a baseline, we use the previously described dual encoder with attention (Figure~\ref{fig:lstm}, right). The inputs are the word embeddings of the original document and of the machine translation. In the output layer, instead of predicting each editor's logit, we predict the logarithm of the post-editing time per source word, following \citet{logtimeprediction2013cohn}. 
We use mean squared error as the loss. 
For our proposed model, we augment this baseline by providing a ``dynamic'' representation of the human post-editor as described below. 


\paragraph{Dynamic editor representations.} In order to obtain an editor's embedding in a real-time setting we do the following: For each new translation session, we store its corresponding embedding, keeping a maximum of 10 previous translations per editor. Whenever an editor's embedding is required, we compute the average of his stored translations into a single vector. This allows updating the editors' representations incrementally in a dynamic fashion, coping with the fact that editors change their behaviour over time as they learn to use the translation interface. 

To introduce a translator vector $\bm{h}$ into the previously described baseline, we increase the input size of the feed-forward layer which receives both encoders' outputs, and we introduce $\bm{h}$ in this step by concatenating it to the encoders' outputs.


\paragraph{Results.}
Both models are evaluated using Pearson correlation between the predicted and real log-times. Results in Table~\ref{table:time_prediction} confirm our hypothesis that {\bf editor representations can be very effective for predicting human post-editing time}, with consistent gains in Pearson correlation ($+$30.11\% in En-Fr and $+$15.05\% in En-De) over the baseline that does not use any editor information. Our approach also allows for initializing and updating editor embeddings dynamically, i.e. without having to retrain the time-prediction model.%
\footnote{This experiment also reveals that previous work on translation quality estimation \citep{Specia2018} using time predictions can have biased results if different types of translators edit different documents. Our editor representations can be potentially useful for removing this bias.}


\begin{table}[]
\small
\begin{tabular}{llcc}
\toprule
   &      & \begin{tabular}[c]{@{}c@{}}Using source text\\ and MT (\%)\end{tabular} & \begin{tabular}[c]{@{}c@{}}Adding dynamic\\ editor embedding (\%)\end{tabular} \\
\midrule
En-Fr & dev  & 19.53                                                                            & {\bf 42.98}                         \\
   & test & 17.58                                                                             & {\bf 47.69}                         \\
En-De & dev  & 27.62                                                                             & {\bf 47.40}                         \\
   & test & 23.67                                                                             & {\bf 38.72}        \\                
\bottomrule
\end{tabular}
\caption{Pearson correlation between real and predicted logarithm of time per word in source text.}
\label{table:time_prediction}
\end{table}


\section{Related Work}\label{sec:related}

There is a long string of work studying the cognitive effort in post-editing machine translation. One of the earliest instances is  
\citet{OBrien2006}, who investigates the relationship between pauses and cognitive effort in post-editing. 
This correlation has also been studied by examination of keystroke logs \citep{LaCruz2012,LaCruz2014}. 
Our results further confirm this, and also identify other characteristics as a fingerprint of the editors: mouse information and jumps. 

More recently, \citet{novice_pro_editors2015moorkens} compare novice and professional post-editors in terms of their suitability as research participants when testing new features of post-editing environments. They conclude that professionals are more efficient but less flexible to interface changes, which confirms the existence of several editor profiles, not necessarily ones better than the others.

Other small-scale studies identify editor behaviour during translation. \cite{asadi2005shortcuts} distinguish between translators who plan ahead and those who type as they think. \cite{daems2019interactive} identify personal preferences between usage of mouse vs. keyboard. \cite{de2013translating} studies differences and similarities in editor behaviour for two language pairs, regarding types of edits, keyboard vs. mouse usage and Web searches.




\citet{Carl2011} have shown that human translators are more productive and accurate when
post-editing MT output than when translating from scratch. This has recently been confirmed by \citet{post_editing_a_novel2018_toral}, who have shown further gains with neural MT compared to phrase-based MT.
\citet{Koponen2012} show HTER \citep{Snover2006} is limited to measure cognitive effort, and suggest post-editing time instead. On the other hand, \citet{herbig2019multi} measure cognitive effort subjectively by directly inquiring translators, and then use a combination of features to predict this cognitive effort -- such task could potentially be improved by including translator representations as an additional feature. \citet{Blain2011} take a more qualitative approach to understanding post-editing
by introducing a measure based on post-editing actions. 
\citet{Specia2011} attempts to predict the post-editing time using quality estimation, and \citet{Koehn2014,SanchezTorron2016} study the impact of machine translation quality in post-editor productivity. 
\citet{Tatsumi2012} study the effect of crowd-sourced post-editing of machine translation output, finding that larger pools of non-experts can frequently produce accurate translations as quickly as experts. 
\citet{Aziz2012} developed a tool for post-editing and assessing machine translation which records data such as editing time, keystrokes,
and translator assessments. A similar tool has been developed by \citet{Denkowski2012,Denkowski2014b}, which is able to learn from post-editing with model adaptation \citep{Denkowski2014}. Our encouraging results on time prediction using editor representations suggests that these representations may also be useful for learning personalized translation models.

\citet{represent_edits_yin19iclr} learn representations of single edits, and include a downstream task: applying these edits to unseen sentences. Wikipedia edits have been studied by \citet{yang2017identifying_semantic_intent} and \citet{faruqui2018wiki_edits}. The latter study what can be learned about language by observing the editing process that cannot be readily learned by observing only raw text. Likewise, we study what can be learned about the translation process by observing how humans type, which cannot be readily learned by observing only the initial and final text.

Our work makes a bridge between the earliest studies on the cognitive effort of human post-editors and modern representation learning techniques, towards embedding human translators on a vector space. We draw inspiration on techniques for learning 
distributed word representations \citep{Mikolov2013,Pennington2014}, which have also been extended for learning user representations for recommendation systems \citep{yahoo_ecommerce_inbox_Grbovic_2015, yahoo_queries_ads_grbovic2016}. 
These techniques usually obtain high-quality embeddings by tuning the system for an auxiliary task, such as predicting a word given its context. 
In our case, we take {\bf editor identification} as the auxiliary task, given a sequence of keytrokes as input. 
A related problem (but with a completely different goal) is the use of keystroke dynamics for user authentication \citep{Monrose2000,Banerjee2012,Kim2018}.  
Unlike this literature, our paper is focused on post-editing of machine-translated text. 
This is more similar to \citet{translator_stylometry_elfiqi2019}, who focus on identifying the translator of a book from his translation style. 
However, we are not interested in the problem of editor identification per se, but only as a means to obtain good representations.

\section{Conclusions}

We introduced and analyzed the largest public dataset so far containing post-editing information retrieved from raw keystrokes. We provided strong evidence that these intermediate steps contain precious information unavailable in the initial plus final translated document, by formulating and providing answers to three research questions: (i) that action sequences can be used to perform accurate editor identification; (ii) that they can be used to learn human post-editor vector representations that cluster together similar editors; and (iii) that these representations help downstream tasks, such as predicting post-editing time. 
In sum, we showed that fine-grained post-editing information is a rich and untapped source of information, and we hope that the dataset we release can foster further research in this area. 

\section*{Acknowledgments}

We  would  like to thank Carla Parra, Alon Lavie, Ricardo Rei, Ant\'onio Lopes, and the anonymous reviewers for their insightful comments. 
This work was partially supported by the EU/FEDER programme under PT2020 (contracts 027767 and 038510)  
and by  the  European  Research  Council  (ERC  StG  DeepSPIN  758969).

\bibliography{mtsummit}
\bibliographystyle{acl_natbib}

\end{document}